\PassOptionsToPackage{numbers}{natbib}
\PassOptionsToPackage{sort}{natbib}

\documentclass{article}

    \usepackage[final]{neurips_2021}

\usepackage[utf8]{inputenc} 
\usepackage[T1]{fontenc}    
\usepackage[colorlinks=true,citecolor=blue]{hyperref}       
\usepackage{url}            
\usepackage{booktabs}       
\usepackage{amsfonts}       
\usepackage{nicefrac}       
\usepackage{microtype}      
\usepackage{xcolor}         
\usepackage{graphicx}
\usepackage{subfigure}
\usepackage{xspace}
\usepackage{enumitem}
\usepackage{amsfonts}
\usepackage{amsmath}
\usepackage[noabbrev,capitalize]{cleveref}
\usepackage{algorithm}
\usepackage{algorithmic}

\newcommand{\ie}{\textit{i.e.,}\xspace}
\newcommand{\eg}{\textit{e.g.,}\xspace}

\newcommand{\fedrecon}{\textsc{FedRecon}\xspace}
\newcommand{\reconeval}{\textsc{ReconEval}\xspace}
\newcommand{\standardeval}{\textsc{StandardEval}\xspace}
\newcommand{\fedavg}{\textsc{FedAvg}\xspace}
\newcommand{\reptile}{\textsc{Reptile}\xspace}
\newcommand{\maml}{\textsc{MAML}\xspace}

\newcommand{\dpsgd}{\textsc{DP-SGD}\xspace}
\newcommand{\finetuning}{\textsc{Finetuning}\xspace}
\newcommand{\fedyogi}{\textsc{FedYogi}\xspace}

\DeclareMathOperator{\E}{\mathbb{E}}
\newcommand{\obj}{F}
\newcommand{\clientObj}{f_i}
\newcommand{\R}{\ensuremath{\mathbb{R}}}
\newcommand{\clientDist}{\ensuremath{\mathcal{P}}}

\newcommand{\vx}{\mathbf{x}}
\newcommand{\lr}{\eta}

\newcommand{\sample}{\xi}
\newcommand{\data}{\ensuremath{\mathcal{D}}}

\title{Federated Reconstruction:\\Partially Local Federated Learning}

\author{%
  Karan Singhal \\
  Google Research \\
  \texttt{karansinghal@google.com} \\
   \And
  Hakim Sidahmed \\
  Google Research \\
  \texttt{hsidahmed@google.com} \\
   \And
  Zachary Garrett \\
  Google Research \\
  \texttt{zachgarrett@google.com} \\
   \And
  Shanshan Wu \\
  Google Research \\
  \texttt{shanshanw@google.com} \\
   \And
  Keith Rush \\
  Google Research \\
  \texttt{krush@google.com} \\
   \And
  Sushant Prakash \\
  Google Research \\
  \texttt{sush@google.com} \\
}

\begin{document}

\maketitle

\begin{abstract}
  Personalization methods in federated learning aim to balance the benefits of federated and local training for data availability, communication cost, and robustness to client heterogeneity. Approaches that require clients to communicate all model parameters can be undesirable due to privacy and communication constraints. Other approaches require always-available or stateful clients, impractical in large-scale cross-device settings. We introduce Federated Reconstruction, the first model-agnostic framework for partially local federated learning suitable for training and inference at scale. We motivate the framework via a connection to model-agnostic meta learning, empirically demonstrate its performance over existing approaches for collaborative filtering and next word prediction, and release an open-source library for evaluating approaches in this setting. We also describe the successful deployment of this approach at scale for federated collaborative filtering in a mobile keyboard application.
\end{abstract}

\section{Introduction}
\label{sec:introduction}
Federated learning is a machine learning setting in which distributed clients solve a learning objective on sensitive data via communication with a coordinating server \citep{mcmahan2017communication}. Typically, clients collaborate to train a single global model under an objective that combines heterogeneous local client objectives. For example, clients may collaborate to train a next word prediction model for a mobile keyboard application without sharing sensitive typing data with other clients or a centralized server \citep{hard2018federated}. This paradigm has been scaled to production and deployed in cross-device settings  \citep{hard2018federated,yang2018applied,apple19wwdc} and cross-silo settings \citep{ClaraTraining,docai}.

However, training a fully global federated model may not always be ideal due to heterogeneity in clients' data distributions. \citet{yu2020salvaging} show that global models can perform worse than purely local (non-federated) models for many clients (\eg those with many training examples). Moreover, in some settings privacy constraints completely prohibit fully global federated training. For instance, for models with user-specific embeddings, such as matrix factorization models for collaborative filtering \citep{koren2009matrix}, naively training a global federated model involves sending updates to user embeddings on the server, directly revealing potentially sensitive individual preferences \citep{gao2020privacy,nikolaenko2013privacy}. 

To address this, we explore partially local federated learning. In this setting, models are partitioned into global $g$ and local parameters $l$ such that local parameters never leave client devices. This enables training on sensitive user-specific parameters as in the collaborative filtering setting, and we show it can also improve robustness to client data heterogeneity and communication cost for other settings, since we are effectively interpolating between local and federated training. Previous works have looked at similar settings \citep{arivazhagan2019federated,liang2020think}. Importantly, these approaches cannot realistically be applied at scale in cross-device settings because they assume clients are stateful or always-available: in practice, clients are sampled from an enormous population with unreliable availability, so approaches that rely on repeated sampling of the same stateful clients are impractical (\citet{kairouz2019advances} [Table 1]). Other work has demonstrated that stateful federated algorithms in partial participation regimes can perform worse than stateless algorithms due to the state becoming "stale" \citep{reddi2020adaptive}.  Previous methods also do not enable inference on new clients unseen during training, preventing real-world deployment.

These limitations motivate a new method for partially local federated learning, balancing the benefits of federated aggregation and local training. This approach should be:

\begin{enumerate}[topsep=0pt,itemsep=-1ex,partopsep=0ex,parsep=1ex]
  \item \textit{Model-agnostic:} works with any model.
  \item \textit{Scalable:} compatible with large-scale cross-device training with partial participation.
  \item \textit{Practical for inference:} new clients can perform inference.
  \item \textit{Fast:} clients can quickly adapt local parameters to their personal data.
\end{enumerate}

\begin{figure}[t]
\vskip 0.0in
\centering
    \includegraphics[width=0.5\linewidth]{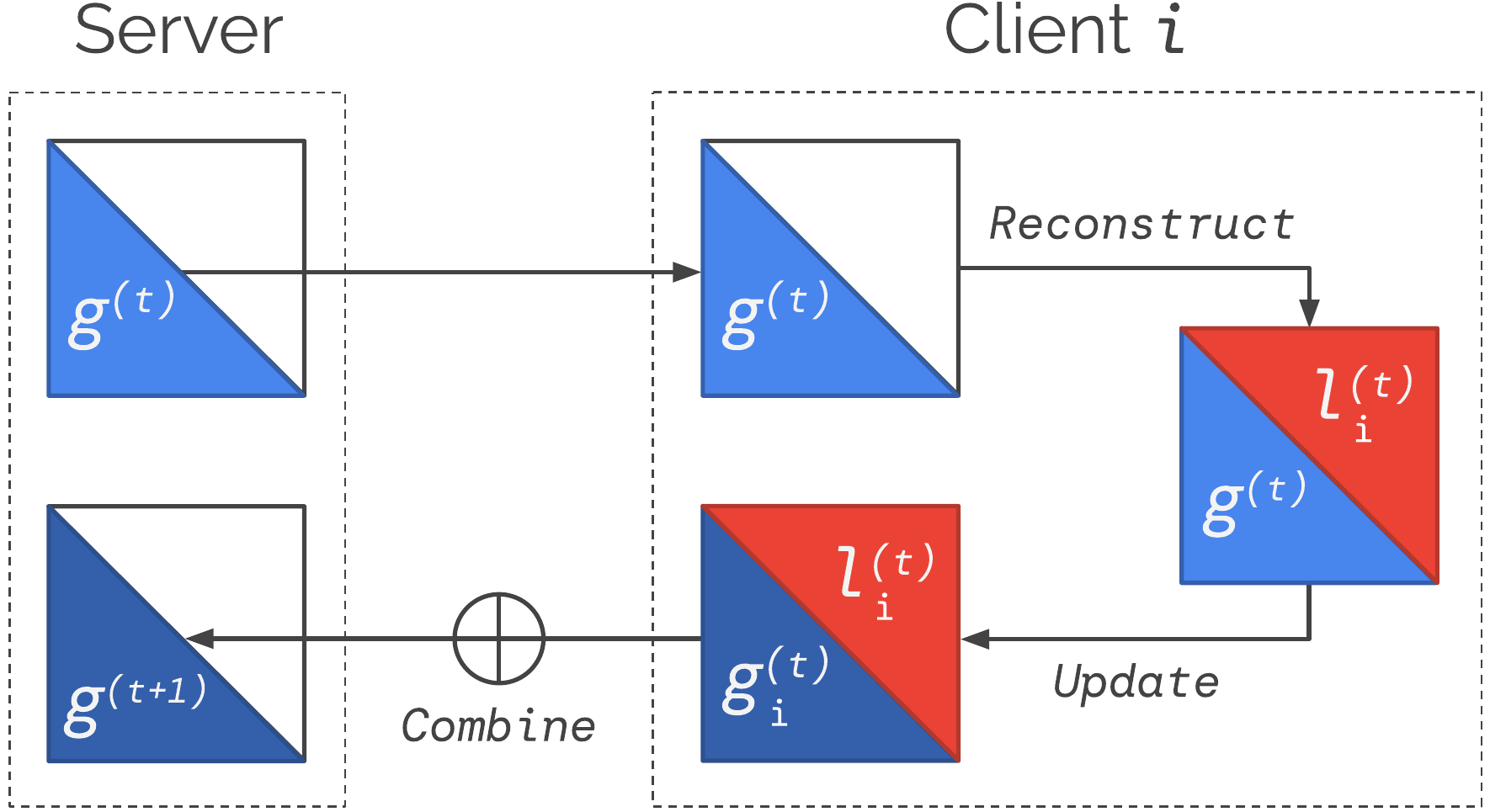}
    \vskip -0.05in
    \caption{Schematic of Federated Reconstruction. Model variables are partitioned into global and local variables. For every round $t$, each participating client $i$ is sent the current global variables, uses them to reconstruct its own local variables, and then updates its copy of the global variables. The server aggregates updates to only the global variables across clients.}
\label{fig:fedrecon_schematic}

\vskip -0.15in
\end{figure}

In this work, we propose combining federated training of global parameters with \emph{reconstruction} of local parameters (see \cref{fig:fedrecon_schematic}). We show that our method relaxes the statefulness requirement of previous work and enables fast personalization for unseen clients without additional communication, even for models without user-specific embeddings.

\textbf{Our contributions:} 
We make the following key contributions:
\begin{itemize}[topsep=0pt,itemsep=-0.5ex,partopsep=0ex,parsep=1ex,leftmargin=5ex]
	\item Introduce a model-agnostic framework for training partially local and partially global models, satisfying the above criteria. We propose a practical algorithm instantiating this framework (\fedrecon).
	\item Justify the algorithm via a connection to model-agnostic meta learning (see \cref{sec:connection_meta_learning}), showing that \fedrecon naturally leads to fast reconstruction at test time (see \cref{tab:movielens_baselines}).
	\item Demonstrate \fedrecon's empirical performance over existing approaches for applications in collaborative filtering and next word prediction, showing that our method outperforms standard centralized and federated training in performance on unseen clients (see \cref{tab:movielens_baselines}), enables fast adaptation to clients' personal data (see \cref{fig:vary_steps}), and matches the performance of other federated personalization techniques with less communication (see \cref{fig:communication_plot}).
	\item Release an open-source library for evaluating algorithms across tasks in this setting.\footnote{ \url{https://github.com/google-research/federated/tree/master/reconstruction} }
	\item Describe the successful deployment of this approach at scale for collaborative filtering in a real-world mobile keyboard application (see \cref{sec:deployment}).
\end{itemize}

\section{Related Work}
\label{sec:related_work}

Previous works have explored personalization of federated models via
finetuning \citep{wang2019federated,yu2020salvaging}, meta learning / bi-level optimization \citep{chen2018federated,jiang2019improving,fallah2020personalized,dinh2020personalized}, and model interpolation \citep{mansour2020three,deng2020adaptive,hanzely2020federated}. Some works aim to improve training convergence with heterogeneous client gradient updates \citep{karimireddy2020scaffold,li2018federated}, while others address client resource heterogeneity \citep{smith2017federated,diao2020heterofl}. All of these approaches require communicating all client parameters during training, which can be unreasonable due to privacy and communication constraints for some models (discussed further in \cref{sec:problem_setting}), which motivates methods that aggregate only part of a model as in our work.

\citet{arivazhagan2019federated} and \citet{liang2020think} aggregate part of a model, but these approaches do not meet the criteria from \cref{sec:introduction}. Similar to other works proposing local parameters \citep{li2021fedbn,huang2021personalized,ge2020fedner}, both approaches require clients to maintain local models across rounds, which is problematic when sampling clients from large populations (criterion 2). \citet{arivazhagan2019federated} assumes that all clients are available for training at all times and do not propose a method for performing inference on new clients (criterion 3). \citet{liang2020think} requires new inference clients to be able to ensemble the outputs of all other clients' local models to evaluate on new data, which is unrealistic in practice due to communication and privacy constraints (criterion 3). These constraints are crucial: with previous methods most clients do not have a practical way to perform inference. Previous methods were also proposed for specific model types (criterion 1): \citet{arivazhagan2019federated} explores personalization layers after shared base layers and \citet{liang2020think} learns personal representations of local data. Finally, as we discuss in \cref{sec:connection_meta_learning}, our method optimizes a meta learning objective for training global parameters that lead to fast reconstruction (criterion 4).

\textbf{Federated Collaborative Filtering: } We evaluate our approach on collaborative filtering \citep{koren2009matrix} in \cref{sec:matrix_factorization_setup}. Prior work has explored federated matrix factorization: \citet{ammad2019federated} avoids sending the user matrix to the server by storing it locally, aggregating only the item matrix globally. \citet{chai2020secure} applies homomorphic encryption to aggregation of the item matrix. \citet{flanagan2020federated} studies federated collaborative filtering as a multi-view learning problem. Each approach requires clients to maintain state, unlike our method. \citet{ammad2019federated} and \citet{chai2020secure} also do not address the problem of inference on unseen users.

\textbf{Federated Meta Learning: } Our approach is motivated by a connection to meta learning, described in \cref{sec:connection_meta_learning}. Other federated learning works have also established connections to meta learning:  \citet{jiang2019improving} observed that training a global federated model that can be easily personalized via finetuning can be studied in the model-agnostic meta learning (\maml) framework \citep{finn2017model}, and \fedavg is performing the distributed version of the \reptile meta learning algorithm presented by \citet{nichol2018first}. \citet{chen2018federated}, \citet{fallah2020personalized}, and \citet{lin2020meta} apply the \maml algorithm and variants in federated settings. \citet{khodak2019adaptive} aims to improve upon these methods by learning client similarities adaptively. These methods do not address the partially local federated learning setting, where some parameters are not aggregated globally.

\section{Partially Local Federated Learning}
\label{sec:problem_setting}

Typically, federated learning of a global model optimizes:
\begin{align} \label{eqn:federated_objective}
\min_{\vx \in \R^d} \obj(\vx) = \E_{i \sim \clientDist}[\clientObj(\vx)]
\end{align}
where $\clientObj(\vx) = \E_{\sample \in \data_i}[\clientObj(\vx, \sample)]$ is the local objective for client $i$, $\vx$ is the $d$-dimensional model parameter vector, $\clientDist$ is the distribution of clients, and $\sample$ is a data sample drawn from client $i$'s data $\data_i$. In practical cross-device settings, $\clientObj(\vx)$ may be highly heterogeneous for different $i$, and the number of available clients may be large and constantly changing due to partial availability. Only a relatively small fraction of clients may be sampled for training.

To motivate partially local federated learning, we begin by considering models that can be partitioned into user-specific parameters and non-user-specific parameters. An example is matrix factorization in the collaborative filtering setting \citep{koren2009matrix,hu2008collaborative}: in this scenario, a ratings matrix $R \in \R^{U \times I}$ representing user preferences is factorized into a user matrix $P \in \R^{U \times K}$ and an items matrix $Q \in \R^{I \times K}$ such that $R \approx PQ^\top$, where $U$ is the number of users and $I$ is the number of items. For each user $u$, this approach yields a $K$-dimensional user-specific embedding $P_u$.

To train this type of model in the federated setting, we cannot naively use the popular \fedavg algorithm \citep{mcmahan2017communication} or other (personalized) algorithms that involving aggregation of all model parameters. A simple application of global learning algorithms might require every client to be sent every other client's personal parameters, which is clearly unreasonable for both privacy and communication. A more sophisticated approach might be to have each client communicate only their own personal parameters with the server. In this case, the server still has access to individual user parameters, which in this setting can be trivially used to recover sensitive user-item affinities, negating the privacy benefit of not centralizing the data (again unreasonable). 

Thus a practical federated learning algorithm for this setting should be \emph{partially local}: it should enable clients to train a subset of parameters entirely on-device. However, approaches that involve stateful clients storing their local parameters across rounds are undesirable in large-scale cross-device settings since clients are unlikely to be sampled repeatedly, causing state to be infrequently available and become stale, degrading performance (\citet{reddi2020adaptive} [Sec. 5.1]). Additionally, since only a fraction of clients participate in training, all other clients will be left without trained local parameters, preventing them from performing inference using the model. In a large population setting with hundreds of millions of clients as described in \cref{sec:deployment}, this can mean 99\%+ of clients do not have a complete model, preventing practical deployment. Thus an algorithm for this setting ideally should not depend on stateful clients and should provide a way to perform inference on unseen clients.

Though we have motivated partially local federated learning via a setting that contains privacy-sensitive user-specific parameters, we will later show that this paradigm can also improve robustness to heterogeneity in $\clientObj(\vx)$ and reduce communication cost, even for models without user-specific parameters. In this case, the partition between local and global parameters is determined by the use-case and communication limitations. As an example, in \cref{sec:nwp_setup} we motivate a next word prediction use-case, where having a partially local model can be useful for handling diverse client inputs while reducing communication.

Achieving partially local federated learning in a practical cross-device setting with large, changing client distribution $\clientDist$ and stateless clients is one of the key contributions of our work.

\section{Federated Reconstruction}

We now introduce the Federated Reconstruction framework. One of the key insights of our approach is that we can relax the requirement for clients to maintain local parameters across rounds by reconstructing local parameters whenever needed, running a reconstruction algorithm $R$ to recover them. Once a client is finished participating in a round, it can discard its reconstructed local parameters. An overview is presented in \cref{fig:fedrecon_schematic}.

\label{sec:federated_reconstruction}
\begin{algorithm}[tb]
   \caption{Federated Reconstruction Training}
   \label{alg:fedrecon_training}

\begin{algorithmic}
\STATE {\bfseries Input:} set of global parameters $\mathcal{G}$, set of local parameters $\mathcal{L}$, dataset split function $S$, reconstruction algorithm $R$, client update algorithm $U$\\
\parbox{.5\linewidth}{

   \STATE {\bfseries Server executes:}
   \begin{ALC@g}
   $g^{(0)} \leftarrow \text{(initialize } \mathcal{G})$
   \FOR{each round $t$}
     
     \item $\mathcal{S}^{(t)} \leftarrow$ (randomly sample $m$ clients)

     \FOR{each client $i \in \mathcal{S}^{(t)}$ \textbf{in parallel}}
       \item $(\Delta_i^{(t)}, n_i) \leftarrow \text{ClientUpdate}(i, g^{(t)})$
     \ENDFOR

   $n = \sum_{i \in \mathcal{S}^{(t)}} n_i$
   
   $g^{(t+1)} \leftarrow g^{(t)} + \lr_s \sum_{i \in \mathcal{S}^{(t)}} \frac{n_i}{n} \Delta_i^{(t)}$
   \ENDFOR
   \end{ALC@g}
   }
   \parbox{.48\linewidth}{
   \STATE {\bfseries ClientUpdate:}
   \begin{ALC@g}
   $(\data_{i,s}, \data_{i,q}) \leftarrow S(\data_i)$
   
   $l_i^{(t)} \leftarrow R(\data_{i,s}, \mathcal{L}, g^{(t)})$
   
   $g_i^{(t)} \leftarrow U(\data_{i,q}, l_i^{(t)}, g^{(t)})$
   
   $\Delta_i^{(t)} \leftarrow g_i^{(t)} - g^{(t)}$
   
   $n_i \leftarrow |\data_{i,q}|$
   
   return $\Delta_i^{(t)}$, $n_i$ to the server
   \end{ALC@g}
   }
\end{algorithmic}

\end{algorithm}

Federated Reconstruction training is presented in \cref{alg:fedrecon_training}. Training proceeds as follows: for each round $t$, the server sends the current global parameters $g^{(t)}$ to each selected client. Selected clients split their local data $\data_i$ into a support set $\data_{i,s}$ and a query set $\data_{i,q}$. Each client uses its support set $\data_{i,s}$ and $g^{(t)}$ as inputs to reconstruction algorithm $R$ to produce its local parameters $l_i^{(t)}$. Then each client then uses its query set $\data_{i, q}$, its local parameters $l_i^{(t)}$, and the global parameters $g^{(t)}$ as inputs to update algorithm $U$ to produce updated global parameters $g_i^{(t)}$. Finally, the server aggregates updates to global parameters across clients. We describe key steps in further detail below.

\textbf{Dataset Split Step:} Clients apply a dataset split function $S$ to their datasets $\data_i$ to produce a support set $\data_{i,s}$ used for reconstruction and a query set $\data_{i,q}$ used for updating global parameters. Typically these sets are disjoint to maximize the meta-generalization ability of the model (see \cref{sec:connection_meta_learning}), but in \cref{sec:appendix_empirical} we show that this assumption may be relaxed if clients don't have sufficient data to partition.

\textbf{Client Reconstruction Step:} Reconstruction of local parameters is performed by algorithm $R$. Though this algorithm can take other forms, in this work we instantiate $R$ as performing $k_r$ local gradient descent steps on initialized local parameters with the global parameters frozen, using the support set $\data_{i,s}$. We show in \cref{sec:connection_meta_learning} this naturally optimizes a well-motivated meta learning objective. Interestingly, this approach is related to gradient-based alternating minimization, a historically successful method for training factored models \citep{jain2013low,gunawardana2005convergence}.

A potential concern with reconstruction is that this may lead to additional client computation cost compared to storing local parameters on clients. However, since clients are unlikely to be reached repeatedly by large-scale cross-device training, in practice this cost is similar to the cost of initializing these local parameters and training them with stateful clients. Additionally, reconstruction provides a natural way for new clients unseen during training to produce their own partially local models offline (see \cref{sec:fed_recon_inference})–without this step, the vast majority of clients would not be able to use the model. Finally, in \cref{sec:connection_meta_learning} we argue and in \cref{sec:experiment_results} we empirically demonstrate that with our approach \textit{just one local gradient descent step} can yield successful reconstruction because global parameters are being trained for fast reconstruction of local parameters.

\textbf{Client Update Step:} Client updates of global parameters are performed by update algorithm $U$. In this work we instantiate $U$ as performing $k_u$ local gradient descent steps on the global parameters, using the query set $\data_{i,q}$. 

\textbf{Server Update Step:} We build on the generalized \fedavg formulation proposed by \citet{reddi2020adaptive}, treating aggregated global parameter updates as an "antigradient" that can be input into different server optimizers (SGD is shown in \cref{alg:fedrecon_training}). Note that the server update operates on a weighted average of client updates as in \citet{mcmahan2017communication}, weighted by $n_i = |\data_{i,q}|$.

We refer to the instantiation of this framework outlined here as \fedrecon below. We address frequently asked questions about \fedrecon and partially local federated learning in \cref{sec:appendix_faq}.

\subsection{Evaluation and Inference}
\label{sec:fed_recon_inference}

To make predictions with global variables $g$ learned using \cref{alg:fedrecon_training}, clients can naturally reconstruct their local models just as they do during training, by using $R$, $g$, and $\data_{i,s}$ to produce local parameters $l$. Then $g$ and $l$ combined make up a fully trained partially local model, which can be evaluated on $\data_{i,q}$. We refer to this evaluation approach as \reconeval below. Note that this can be applied to clients unseen during training (most clients in large-scale settings), enabling inference for these clients.\footnote{In this work we focus on new clients that have some local data for reconstruction; our method can be easily extended to learn a global default for the local parameters. We also show that skipping reconstruction can be reasonable for some tasks in \cref{sec:experiment_results}.}

Reconstruction for inference is performed offline, independently of any federated process, so clients can perform reconstruction once and store local parameters for repeated use, optionally refreshing them periodically if they have new local data.

\subsection{Connection to Meta Learning}
\label{sec:connection_meta_learning}

Our framework is naturally motivated via meta learning. Given that \reconeval involves clients doing (gradient-based) reconstruction using global parameters, we ask: \textit{Can we train global parameters conducive to fast reconstruction of local parameters?} 

We can easily formulate this question in the language of model-agnostic meta learning \citep{finn2017model}. The heterogeneous client distribution $\clientDist$ corresponds to the heterogeneous distribution of tasks; each round (episode) we sample a batch of clients in the hope of meta-generalizing to unseen clients. Each client has a support dataset for reconstruction and a query dataset for global parameter updates. Our meta-parameters are $g$ and our task-specific parameters are $l_i$ for client $i$. We want to find $g$ minimizing the objective:
\begin{align} \label{eqn:meta_objective}
\E_{i \sim \clientDist}\clientObj(g \mathbin\Vert l_i) = \E_{i \sim \clientDist} f_i(g \mathbin\Vert R(\data_{i,s},\mathcal{L}, g)]
\end{align}
where $g \mathbin\Vert l_i$ denotes the concatenation of $g$ and $l_i$ and $\clientObj(g \mathbin\Vert l_i) = \E_{\sample \in \data_{i,q}}[\clientObj(g \mathbin\Vert l_i, \sample)]$. 

In \cref{sec:meta_learning_proof} we show that the instantiation of our framework where $R$ performs $k_r \ge 1$ steps of gradient descent on initialized local parameters using $\data_{i,s}$ and $U$ performs $k_u = 1$ step of gradient descent using $\data_{i,q}$ is \emph{already} minimizing the first-order terms in this objective (\ie this version of \fedrecon is performing first-order meta learning). Intuitively, reconstruction corresponds to the MAML ``inner loop'' and the global parameter update corresponds to the ``outer loop''; we test the same way we train (via reconstruction), a common pattern in meta learning. 

Thus \fedrecon trains global parameters $g$ for fast reconstruction of local parameters $l$, enabling partially local federated learning without requiring clients to maintain state. In \cref{sec:experiment_results} we observe that our method empirically produces $g$ more conducive to fast, performant reconstruction on unseen clients than standard centralized or federated training (\eg see \textsc{Server+ReconEval} vs. \textsc{FedRecon} in \cref{tab:movielens_baselines}). We see in \cref{fig:vary_steps} that \textit{just one reconstruction step} is sufficient to recover the majority of performance.

\section{Experimental Evaluation}

\subsection{Tasks and Methods}
\label{sec:experiment_tasks_methods}
We next describe experiments validating \fedrecon on matrix factorization and next word prediction. We aim to determine whether reconstruction can enable practical partially local federated learning with fast personalization for new clients, including in settings without user-specific embeddings.

\subsubsection{Matrix Factorization}
\label{sec:matrix_factorization_setup}
We evaluate on federated matrix factorization using the popular MovieLens 1M collaborative filtering dataset \citep{harper2015movielens}. We perform two kinds of evaluation:

\begin{enumerate}
    \item \standardeval on \textit{seen} users, those users who participated in at least one round of federated training. We split each user's ratings into 80\% train, 10\% validation, and 10\% test by timestamp. We train on all users' train ratings, and report results on users test ratings.
    \item \reconeval on \textit{unseen} users, those users who did not participate at all during federated training. We split the users randomly into 80\% train, 10\% validation, and 10\% test; we train with the train users and report results on test users.
\end{enumerate}

The model learns $P$ and $Q$ such that $R \approx PQ^{\top}$ as discussed in \cref{sec:problem_setting}, with embedding dimensionality $K=50$. We apply \fedrecon with local user embeddings $P_u$ and global item matrix $Q$. We report root-mean-square-error (RMSE) and rating prediction accuracy. We compare centralized training, \fedavg, and \fedrecon in \cref{tab:movielens_baselines}. See also \cref{sec:appendix_dataset_models_matfac} for more details on the dataset, model, and hyperparameter choices.

\subsubsection{Next Word Prediction}
\label{sec:nwp_setup}
We also aim to determine whether Federated Reconstruction can be successfully applied in settings without user-specific embeddings to improve robustness to client heterogeneity and communication cost, since our approach is agnostic to which parameters are chosen as local/global. We apply \fedrecon to next word prediction because the task provides a natural motivation for personalization: different clients often have highly heterogeneous data, \eg if they use different slang, but language models typically have a fixed vocabulary. We propose improving the ability of a language model to capture diverse inputs using local \emph{out-of-vocabulary} (OOV) embeddings. OOV embeddings are a common application of the hashing trick \citep{weinberger2009feature} in deep learning; combining them with \fedrecon enables language models to effectively allow for personal input vocabularies for different clients. For example, if client $i$ frequently uses OOV token $t_i$ and client $j$ uses OOV token $t_j$, each client's corresponding local OOV embedding can learn to reflect this (even if the OOV embeddings collide). So adding local OOV embeddings with the core global vocabulary fixed can lead to improved personalization without more communication per round; we will also show that we can reduce the size of the core model (reducing communication) and get further benefits.

We perform next word prediction with the federated Stack Overflow dataset introduced in \citet{TFFSO2019}. We use an LSTM model and process data similarly to \citet{reddi2020adaptive}, comparing to their best \fedyogi result. To demonstrate that reconstruction can be used to reduce model size, we describe experiments with vocabulary sizes [1000, 5000, 10,000]. See \cref{sec:appendix_dataset_models_nwp} for details on the dataset, model, and hyperparameter choices.

\subsection{Results and Discussion}
\label{sec:experiment_results}

\begin{table}[t]
\caption{Movielens matrix factorization root-mean-square-error (lower is better) and rating prediction accuracy (higher is better). \standardeval is on seen users, \reconeval is on held-out users. Results within 2\% of best for each metric are in bold.}
\label{tab:movielens_baselines}
\vskip 0.15in
\begin{center}
\begin{small}
\begin{sc}
\begin{tabular}{lccr}
\toprule
 & RMSE $\downarrow$ & Accuracy $\uparrow$  \\
\midrule
Centralized + Standard Eval    & \textbf{.923} & \textbf{43.2} \\
Centralized + ReconEval    & 1.36 & 40.8 \\
FedAvg + Standard Eval & .939 & 41.5 \\
FedAvg + ReconEval   & .934 & 40.0 \\
\fedrecon (Ours)   & \textbf{.907} & \textbf{43.3} \\

\bottomrule
\end{tabular}
\end{sc}
\end{small}
\end{center}
\vskip -0.1in
\end{table}

\begin{table}[t]
\caption{Stack Overflow next word prediction accuracy and communication per round, per client. \fedyogi and \textsc{OOV/Full Finetuning} require communication of all model parameters, \fedrecon does not (see \cref{fig:communication_plot}). Results within 2\% of best for each vocabulary size are in bold.}
\label{tab:so_nwp_baselines}
\begin{center}
\begin{small}
\begin{sc}
\begin{tabular}{lcccr}
\toprule
Vocab. Size & 1K & 5K & 10K & Communication \\
\midrule
FedYogi   & 24.3 & 26.3 & 26.7  & $2|l| + 2|g|$\\
\fedrecon (1 OOV)   & 24.1 & 26.2 & 26.4  & $\mathbf{2|g|}$ \\
\fedrecon (500 OOV)   & 29.6 & 28.1 & 27.7 & $\mathbf{2|g|}$  \\
\midrule
OOV Finetuning (500 OOV)   & 30.0 & 28.1 & 27.9  & $2|l| + 2|g|$ \\
Full Finetuning (500 OOV)   & \textbf{30.8} & \textbf{29.2} & \textbf{28.8}  & $2|l| + 2|g|$ \\
FedRecon+Finetune (500 OOV)   & \textbf{30.7} & \textbf{28.9} & \textbf{28.6}  &  $\mathbf{2|g|}$ \\

\bottomrule
\end{tabular}
\end{sc}
\end{small}
\end{center}
\vskip -0.2in
\end{table}

In \cref{tab:movielens_baselines,tab:so_nwp_baselines} we present results for matrix factorization and next word prediction for \fedrecon and baselines. We call out several key comparisons below; more results can be found in \cref{sec:appendix_empirical}. 

For the MovieLens task \fedrecon is able to match the performance of \textsc{Centralized + Standard Eval} despite performing a more difficult task: as described in \cref{sec:matrix_factorization_setup}, \fedrecon is using \reconeval to evaluate on \textit{held-out users}, reconstructing user embeddings for them and then evaluating. As is typical for server-trained matrix factorization models, \textsc{Centralized + Standard Eval} is only being evaluated on \textit{held-out ratings for seen users}. Note that we would not be able to evaluate on unseen users since they do not have trained user embeddings (randomly initializing them produces garbage results). If we reconstruct user embeddings for unseen users and then evaluate as in \textsc{Centralized + ReconEval} (we argue this is a fairer comparison with \fedrecon), we see that performance is significantly worse than \fedrecon and server-evaluation on seen users. One interesting finding was that the results of this seemed to vary widely across different users, with some users reconstructing embeddings no better than random initialization, while most others reconstructed better embeddings.\footnote{For this experiment, we repeat 500 times: sample 50 clients each time and perform \reconeval, reporting average metrics. Across runs, we observe large standard deviations of 1.7\% accuracy (absolute) and 0.53 RMSE.} We see a similar result with \fedavg for the MovieLens task, where \fedavg with standard evaluation on seen users\footnote{Note that for this task, \fedavg is equivalent in result to the stateful \textsc{FedPer} approach in \citet{arivazhagan2019federated}, since each client only updates its own user embedding. The user embeddings are stored on the server here, but this does not affect the result. Performance reduction may be caused by user embeddings getting "stale" across rounds which may occur when stateful algorithms are applied in cross-device FL, see \cref{sec:appendix_faq}.} performs a bit worse than \textsc{Centralized + Standard Eval}, and performance for \reconeval on unseen users is significantly worse than \fedrecon. This indicates that \fedrecon is doing a better job of training global parameters \textit{so they can reconstruct local parameters} than other approaches, as motivated in \cref{sec:connection_meta_learning}.  Moreover, \fedrecon is doing this despite not having direct access to the data or the user-specific parameters–enabling this approach in settings where centralized training or \fedavg is impossible. 

\begin{figure}[t]
\vskip 0.0in
\centering
    \includegraphics[width=0.5\linewidth]{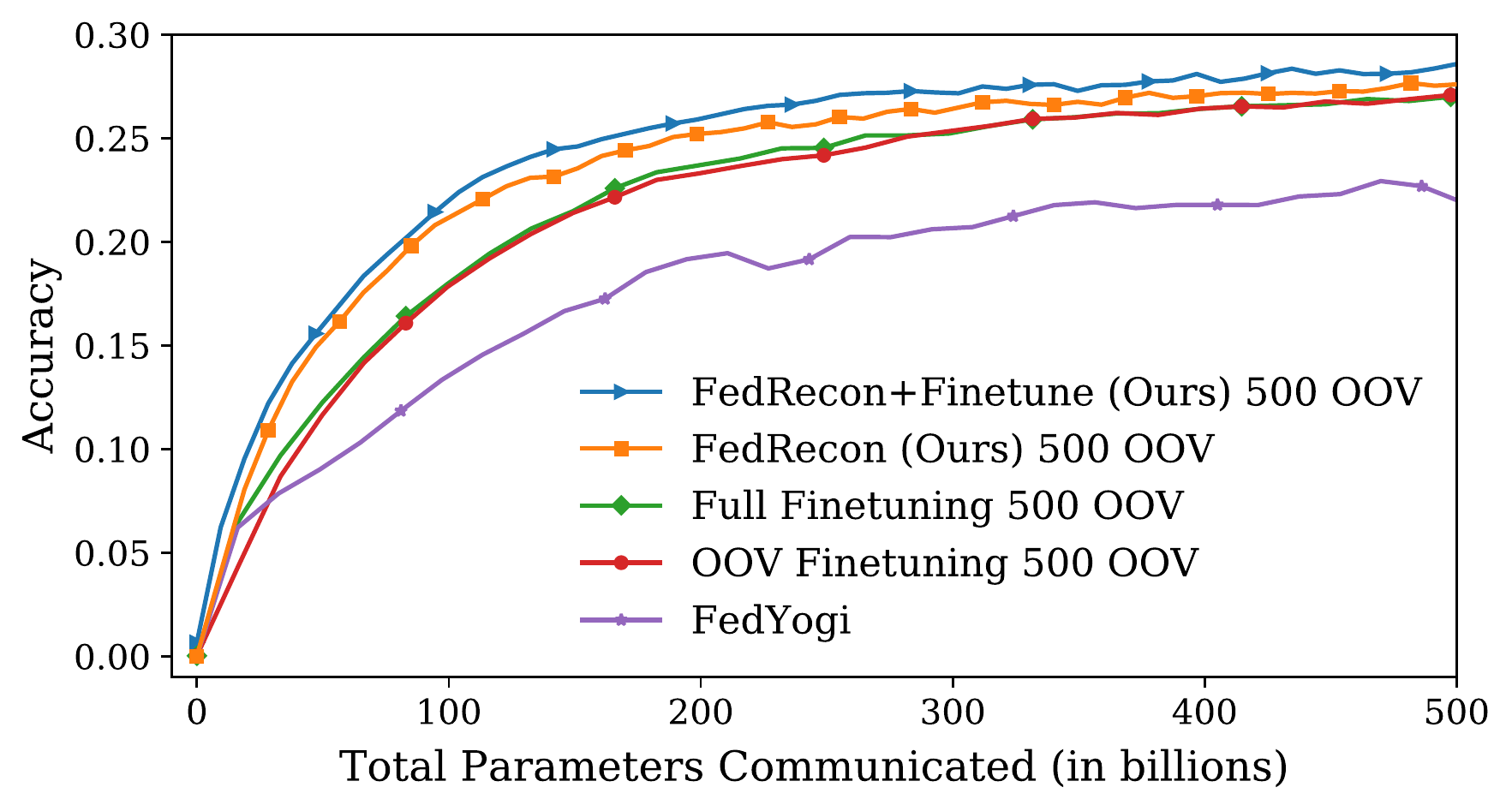}
    \vskip -0.1in
    \caption{Accuracy as a function of total parameters communicated across all clients for \fedrecon and baselines for Stack Overflow next word prediction.}
\label{fig:communication_plot}

\vskip -0.05in
\end{figure}

In the first section of the Stack Overflow results in \cref{tab:so_nwp_baselines}, we compare \textsc{FedYogi} (an adaptive variant of \fedavg introduced by \citet{reddi2020adaptive}) with \fedrecon, showing that enabling \fedrecon with 500 local OOV embeddings significantly boosts accuracy for every vocabulary size. Interestingly, we observe that accuracy actually improves for smaller vocabulary sizes for \textsc{FedRecon (500 OOV)}, whereas the reverse holds for \fedyogi and \textsc{FedRecon (1 OOV)}. We posit that this is because decreasing the vocabulary size effectively increases the amount of "training data" available for the local part of the model, since OOV embeddings are only used (and trained) when tokens are out-of-vocabulary; this is useful only when the local part of the model has sufficient capacity via the number of OOV embeddings. This hypothesis is consistent with vocabulary coverage: a 10K vocabulary covers 86.9\% of the tokens in the dataset, a 5K vocabulary covers 80.1\%, and a 1K vocabulary covers 49.2\%; we see that difference in results for \textsc{FedRecon (500 OOV)} is greater between 1K and 5K than between 5K and 10K. We caution that reducing vocabulary size may be undesirable in some cases: reducing the size of the vocabulary also restricts the output tokens of the model.

\begin{figure}[t]
\vskip 0.0in
\centering
    \includegraphics[width=.9\columnwidth]{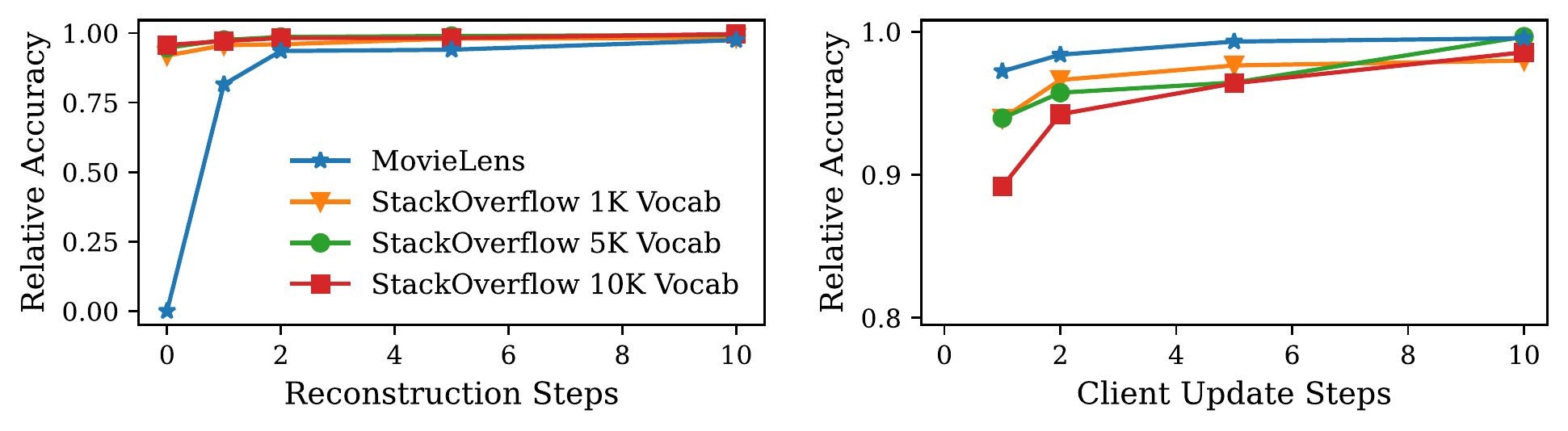}
    \vskip -0.1in
    \caption{Accuracy compared to base \fedrecon when varying the number of reconstruction steps for local parameters (left plot) and client update steps for global parameters (right plot).}
\label{fig:vary_steps}

\vskip -0.15in
\end{figure}

\textbf{Comparing with Finetuning: } In \cref{tab:so_nwp_baselines} we compare \fedrecon with \finetuning \citep{wang2019federated,yu2020salvaging} to study whether reconstruction can provide similar benefits as global personalization methods. In our implementation we train a fully global model using \fedyogi, perform local gradient steps to finetune part of the model using the support set, and then evaluate on the query set (same sets as used for \fedrecon). For \textsc{OOV Finetuning}, the OOV parameters only are finetuned using the support set (comparable to \fedrecon), and for \textsc{Full Finetuning} all parameters are finetuned. Comparing \textsc{FedRecon (500 OOV)} and \textsc{OOV Finetuning}, we see that reconstructing local embeddings performs similarly to finetuning pre-trained OOV embeddings, despite \fedrecon not communicating the local parameters $l$ to the server. \textsc{Full Finetuning} from \cref{tab:so_nwp_baselines} achieves better accuracy since all parameters are finetuned. To compare this fairly with reconstruction, we perform \textsc{FedRecon+Finetune}, where the support set is used first to reconstruct local parameters and then to finetune global parameters before evaluation. We also see that we can get comparable results, indicating that reconstruction can enable personalization on (potentially privacy-sensitive) local parameters while reducing communication. See \cref{fig:communication_plot} for a comparison of different approaches by the total number of parameters communicated–we see an advantage for \fedrecon, particularly for lower total communication.

\textbf{Varying Reconstruction Steps: } In \cref{sec:connection_meta_learning} we described a connection between our framework and \maml \citep{finn2017model}, which has been a successful paradigm for fast adaptation to tasks with few steps. In \cref{fig:vary_steps} we perform \fedrecon for varying numbers of reconstruction steps $k_r \in [0, 1, 2, 5, 10]$ and plot the accuracy as a fraction of accuracy across tasks from \cref{tab:movielens_baselines,tab:so_nwp_baselines}. We see that for zero reconstruction steps (an ablation skipping reconstruction), MovieLens accuracy is 0.0, as expected (all user embeddings are randomly initialized). Relative accuracy for Stack Overflow NWP settings remains above 90\%, suggesting that for this task clients can still perform inference with a \fedrecon-trained model even without any data to reconstruct. Importantly, \textit{just one reconstruction step} is required to recover the majority of remaining performance across both tasks, indicating that \fedrecon learns global parameters conducive to fast reconstruction.

\textbf{Varying Client Update Steps: } In \cref{sec:connection_meta_learning} we showed that gradient-based \fedrecon, involving $k_r \geq 1$ reconstruction steps and $k_u = 1$ client update steps, is minimizing a first-order meta learning objective for training global parameters that yield good reconstructions. In \cref{fig:vary_steps} we perform \fedrecon with $k_u \in [1, 2, 5, 10]$ and compute relative accuracy as a fraction of accuracy across tasks from \cref{tab:movielens_baselines,tab:so_nwp_baselines}. For each experiment we run for a fixed number of rounds. We see that 1 step recovers almost all of the accuracy and adding more steps gradually increases accuracy further. Interestingly, we observe that for $k_u = 1$ training proceeds significantly slower than for other values such that performance is still slightly increasing after the fixed number of rounds. This is analogous to the difference between \fedavg and \textsc{FedSGD} \citep{mcmahan2017communication}. While \textsc{FedSGD} is optimizing the original learning objective, \fedavg often achieves similar performance in significantly fewer rounds by adding multiple gradient steps on aggregated parameters.

We present further baselines and ablations in \cref{sec:appendix_empirical}.

\section{Open-Source Library}
\label{sec:open_source_library}

We are releasing a code framework for expressing and evaluating practical partially local federated models built on the popular TensorFlow Federated library \citep{TFF}. The code is released under Apache License 2.0. In addition to allowing for easy reproduction of our experiments, the framework provides a flexible, well-documented interface for researchers and modelers to run simulations in this setting with models and tasks of their choice. Users can take any existing Keras model and plug it into this framework with just a few lines of code. We provide libraries for training and evaluation for MovieLens matrix factorization and Stack Overflow next word prediction, which can be easily extended for new tasks. We hope that the release of this framework spurs further research and lowers the barrier to more practical applications.

\section{Deployment in a Mobile Keyboard Application}
\label{sec:deployment}

A key differentiator of our method is that it scales to practical training and inference in cross-device settings with large populations. To validate this, we deployed \fedrecon to a mobile keyboard application with hundreds of millions of federated learning clients. We used a system similar to \citet{bonawitz2019towards} to deploy \fedrecon for training. Note that the system does not support stateful clients given the issues with large-scale stateful training described in \cref{sec:federated_reconstruction}, so a stateless approach was necessary for deployment.

Users of the mobile keyboard application often use \emph{expressions} (GIFs, stickers) to communicate with others in \eg chat applications. Different users are highly heterogeneous in the style of expressions they use, which makes the problem a natural fit for collaborative filtering to predict new expressions a user might want to share. We trained matrix factorization models as described in \cref{sec:matrix_factorization_setup}, where the number of items ranged from hundreds to tens of thousands depending on the type of expression.

Training in production brought challenges due to data sparsity. Depending on the task, some clients had very few examples, if \eg they didn't commonly share stickers via the keyboard application. To ensure clients with just one example weren't just adding noise to the training process by participating, we oversampled clients and filtered out the contributions of clients without at least some number of examples. We reused examples between the support and query sets as described in \cref{sec:appendix_empirical} to ensure all examples were used for both reconstruction and global updates.

Another practical challenge we faced was orthogonal to our method and commonly faced in real-world federated learning applications: heterogeneity in client resources and availability meant that some participating clients would drop out before sending updates to the server. We found that the simple strategy of oversampling clients and neglecting updates from dropped-out clients appeared to perform well, but we believe studying the fairness implications of this is a valuable area for future work.

After successful training, the resulting model was deployed for inference in predicting potential new expressions a user might share, which led to an increase of 29.3\% in click-through-rate for expression recommendations. We hope that this successful deployment of \fedrecon demonstrates the practicality of our approach and leads the way for further real-world applications.

\section{Conclusion}
We introduced Federated Reconstruction, a model-agnostic framework for fast partially local federated learning suitable for training and inference at scale. We justified \fedrecon via a connection to meta learning and empirically validated the algorithm for collaborative filtering and next message prediction, showing that it can improve performance on unseen clients and enable fast personalization with less communication. We also released an open-source library for partially local federated learning and described a successful production deployment. Future work may explore the optimal balance of local and global parameters and the application of differential privacy to global parameters (see \cref{sec:appendix_privacy_implications}).

\begin{ack}
We thank Brendan McMahan, Lin Ning, Zachary Charles, Warren Morningstar, Daniel Ramage, Jakub Kone\v{c}n\'{y}, Blaise Ag\"uera y Arcas, and Jay Yagnik from Google Research for their helpful comments and discussions. We also thank Wei Li, Matt Newton, and Yang Lu for their collaboration towards deployment.
\end{ack}

\bibliographystyle{ACM-Reference-Format}
\bibliography{fedrecon}

\newpage
\appendix

\section{Frequently Asked Questions}
\label{sec:appendix_faq}

$>$ Why can't clients just remember their local parameters? Why can't we use a stateful algorithm?

In real-world cross-device FL settings, the population size is huge (\eg millions of clients), only a small number of clients participate in each round (\eg 200 clients), and a client usually participates \textbf{at most once} during the entire training process. Under this setting, algorithms cannot rely on client state, such as local parameters, from a previous round because in almost every case it will not exist. In fact, stateful algorithms result in performance degradation in the cross-device setting due to state getting "stale" between rounds (see the \textsc{SCAFFOLD} discussion in \citet{reddi2020adaptive} Sec. 5.1). We empirically observe a performance difference in \cref{tab:movielens_baselines}, where \textsc{FedAvg} (which performs identically to the stateful \textsc{FedPer} approach of \citet{arivazhagan2019federated} for this task) does not match the performance of \fedrecon for matrix factorization. Additionally, in partially local FL, a stateful algorithm would result in all non-sampled clients being without trained local parameters. In the real-world setting from \cref{sec:deployment}, this would mean more than \textbf{99\%} of clients would not have working models, preventing practical deployment. Note that clients can still optionally store final local variables for inference after training as mentioned in \cref{sec:fed_recon_inference}, since this is independent of any federated process.

$>$ Why perform reconstruction of local parameters? Isn't reconstruction wasteful?

Since clients are very unlikely to be reached repeatedly by large-scale cross-device training (see above answer), in practice this cost is nearly the same as the cost of initializing these local parameters and training them with stateful clients. Additionally, reconstruction provides a natural way for new clients unseen during training to produce their own partially local models offline (see \cref{sec:fed_recon_inference})–without this step, the vast majority of clients in the real-world deployment described in \cref{sec:deployment} would not be able to use the model. Reconstruction also ensures local parameters are always fresh, avoiding staleness issues (see above answer). Reconstruction also allows us to save on communication cost compared to fully global models, which is typically more of a limiting resource in federated learning than local client computation \citep{kairouz2019advances,li2020federated}. See the improvement in the accuracy-communication tradeoff in \cref{tab:so_nwp_baselines,fig:communication_plot}. Lastly, in \cref{sec:connection_meta_learning,sec:meta_learning_proof} we provide intuitive and theoretical arguments for why reconstruction naturally leads to training global parameters that can be easily used to train personal local parameters, similar to the existing intuition around Model-Agnostic Meta Learning methods \citep{finn2017model}. In \cref{fig:vary_steps,tab:movielens_baselines}, we also empirically demonstrate that reconstruction leads to performant final models with minimal gradient steps (even on unseen clients).

$>$ How is this different from \textit{Paper X}?

Our work differs from other work in that (1) it addresses partially local federated learning and (2) it proposes a stateless algorithm for this setting, making partially local federated learning practical in large-scale cross-device settings. Many other federated learning works either don't address (1) \eg because they do not involve models where some (privacy-sensitive) parameters are local and some parameters are global, or don't address (2) \eg because they introduce algorithms that require stateful clients. Our method has also been extensively validated through simulation experiments using Stack Overflow and MovieLens user data and a real-world deployment to a cross-device setting with hundreds of millions of clients (rare in FL papers). We have also open-sourced the code that was deployed, making practical partially local federated learning widely available; we know of no other work that has done this. We discuss specific differences from previous works in \cref{sec:related_work}. 

$>$ How do we know the approach converges?

We show that our approach converges empirically in \cref{fig:movielens_optimizers,fig:nwp_losses,fig:communication_plot}. We also show in \cref{tab:movielens_baselines} that we achieve better results on unseen users than a \textbf{server-trained} matrix factorization model, again demonstrating empirical performance. We also show a theoretical connection to MAML in \cref{sec:meta_learning_proof}, which justifies the fast reconstruction in \cref{fig:vary_steps}. While we do not present additional theoretical convergence results, one of our core contributions in this work is our deployment of this approach at scale in a real-world setting with hundreds of millions of clients, improving recommendation CTR by 29.3\% (see \cref{sec:deployment}). We believe that this contribution does more to validate the approach than theoretical convergence results. 

$>$ How do we split the model into global and local parts in partially local FL?

If the model has user-specific or very privacy-sensitive parameters (as in the matrix factorization case), then these parameters can naturally be local parameters. For other models, the global-local split follows from the use-case and communication requirements. If communication is a concern, making more variables local may be a way of reducing communication cost. As an example, in \cref{sec:nwp_setup} we motivate the next word prediction use-case, where having a partially local model with local OOV embeddings can be useful for handling diverse client inputs while reducing communication.

\section{Proof of Connection to Meta Learning}
\label{sec:meta_learning_proof}
As described in \cref{sec:connection_meta_learning}, our meta-parameters are $g$ and our task-specific parameters are $l_i$ for client $i$. We want to find $g$ minimizing the objective:
\begin{align} \label{eqn:meta_objective}
\E_{i \sim \clientDist}\clientObj(g \mathbin\Vert l_i) = \E_{i \sim \clientDist} f_i(g \mathbin\Vert R(\data_{i,s},\mathcal{L}, g)]
\end{align}
where $g \mathbin\Vert l_i$ denotes the concatenation of $g$ and $l_i$ and $\clientObj(g \mathbin\Vert l_i) = \E_{\sample \in \data_{i,q}}[\clientObj(g \mathbin\Vert l_i, \sample)]$. 

We now show that the instantiation of our framework where $R$ performs $k_r \ge 1$ steps of gradient descent on initialized local parameters using $\data_{i,s}$ and $U$ performs $k_u = 1$ step of gradient descent using $\data_{i,q}$ is \emph{already} minimizing the first-order terms in this objective (\ie this version of \fedrecon is performing first-order meta learning).

Taking the server update from \cref{alg:fedrecon_training} for round $t$, we have:
\begin{align*}
g^{(t+1)} \leftarrow g^{(t)} + \lr_s \sum_{i \in \mathcal{S}^{(t)}} \frac{n_i}{n} \Delta_i^{(t)}
\end{align*}
Given that $U$ performs $k_u = 1$ step of SGD with learning rate $\lr_u$, the right side is equivalent to:
\begin{align*}
g^{(t)} + \lr_s \sum_{i \in \mathcal{S}^{(t)}} \frac{n_i}{n} ((g^{(t)} - \lr_u \nabla_{g} \clientObj(g^{(t)} \mathbin\Vert l_i^{(t)})) - g^{(t)})
\end{align*}
Simplifying and writing $l_i^{(t)}$ in terms of $R$:
\begin{align*}
g^{(t)} - \lr_s \lr_u \sum_{i \in \mathcal{S}^{(t)}} \frac{n_i}{n} (\nabla_{g} \clientObj(g^{(t)} \mathbin\Vert R(\data_{i,s}, \mathcal{L}, g^{(t)})))
\end{align*}
Note that this corresponds to sampling a batch of tasks and performing SGD on the global parameters in \cref{eqn:meta_objective} with learning rate $\lr_s \lr_u$, weighted by the number of examples by task (this can be omitted if desired). An important detail is that the result of $R$ is treated as a constant in computing $\nabla_{g}
\clientObj$. We argue that this corresponds to neglecting the second-order terms in the gradient. Since $R$ performs $k_r$ steps of SGD, we can write out each step:
\begin{align*}
l_i^{(t,1)} &= l_i^{(t,0)} - \lr_r \nabla_{l_i} \clientObj(g^{(t)} \mathbin\Vert l_i^{(t,0)})\\
l_i^{(t,2)} &= l_i^{(t,1)} - \lr_r \nabla_{l_i} \clientObj(g^{(t)} \mathbin\Vert l_i^{(t,1)})\\
&\ldots \\ 
l_i^{(t)} = l_i^{(t,k_r)} &= l_i^{(t,k_r-1)} - \lr_r \nabla_{l_i} \clientObj(g^{(t)} \mathbin\Vert l_i^{(t,k_r-1)})\\
\end{align*}
where $l_i^{(t,0)}$ is a random initialization of the parameters in $\mathcal{L}$. So using the chain rule $\nabla_{g} \clientObj(g^{(t)} \mathbin\Vert R(\data_{i,s}, \mathcal{L}, g^{(t)}))$ can be written as:
{\small
    \begin{align*}
        &\nabla_{g} \clientObj(g^{(t)} \mathbin\Vert l_i^{(t,k_r)}) + \nabla_{l_i} \clientObj(g^{(t)} \mathbin\Vert l_i^{(t,k_r)}) \nabla_{g}\left(l_i^{(t,0)}- \lr_r \sum_{j=0}^{k_r-1}\nabla_{l_i} \clientObj(g^{(t)} \mathbin\Vert l_i^{(t,j)})\right)\\
        &=  \nabla_{g} \clientObj(g^{(t)} \mathbin\Vert l_i^{(t,k_r)}) - \lr_r \nabla_{l_i} \clientObj(g^{(t)} \mathbin\Vert l_i^{(t,k_r)}) \sum_{j=0}^{k_r-1}\nabla_{g} \nabla_{l_i} \clientObj(g^{(t)} \mathbin\Vert l_i^{(t,j)})
    \end{align*}
}%
Treating the output of $R$ as constant in $g$ therefore corresponds to dropping the Jacobian-gradient terms (which are second-order partial derivatives in elements of $g$ and $l_i$), leaving us with $\nabla_{g} \clientObj(g^{(t)} \mathbin\Vert l_i^{(t,k_r)})$. Using the full matrix of these terms would lead to additional computational cost and memory cost quadratic in the number of local parameters, which is generally impractical on heterogeneous client devices. Neglecting similar terms (these terms reduce to the Hessian of $f_i$ in the case that $g = l_i$) has been shown to cause minimal drop in performance of meta learning algorithms \citep{nichol2018first,finn2017model}. 

We have shown \fedrecon trains global parameters $g$ for fast reconstruction of local parameters $l$, enabling partially local federated learning without requiring clients to maintain state. In \cref{sec:experiment_results} we observe that our method empirically produces $g$ more conducive to fast, performant reconstruction on unseen clients than standard centralized or federated training (\eg see \textsc{Server+ReconEval} vs. \textsc{FedRecon} in \cref{tab:movielens_baselines}). We see in \cref{fig:vary_steps} that \textit{just one reconstruction step} is sufficient to recover the majority of performance.

\section{Datasets, Models, and Hyperparameters}
\label{sec:appendix_dataset_models}

Below we provide further detail on our evaluation tasks, including descriptions of datasets, models, and hyperparameters. Our open-source framework described in \cref{sec:open_source_library} contains commands to reproduce these tasks.

\subsection{Matrix Factorization}
\label{sec:appendix_dataset_models_matfac}

We evaluate on federated matrix factorization using the popular MovieLens 1M collaborative filtering dataset \citep{harper2015movielens}. The dataset consists of 1,000,209 ratings on 3,706 movies from 6,040 users who joined MovieLens in 2000.\footnote{Data was provided voluntarily by users. The full dataset includes some demographic information for these users, \eg their gender, but we do not use this information in this work.} The dataset is licensed for research use \citep{movielens_license}.

We perform two kinds of evaluation: (1) standard (federated) evaluation on seen users and (2) \reconeval on \textit{unseen} users. For (1), we split each user's ratings into 80\% train, 10\% validation, and 10\% test by timestamp. This is the typical evaluation setup in the matrix factorization literature \citep{hu2008collaborative}. For (2) we split the users randomly into 80\% train, 10\% validation, and 10\% test; we train with the train users and report results on test users. This setup tests the model's ability to generalize to unseen users without trained user embeddings. Note that without reconstruction of user embeddings, we would expect the model to perform poorly on unseen users with randomly initialized user embeddings, and this is what we observe in \cref{tab:movielens_appendix_results} (compare \textsc{Centralized + StandardEval (Unseen)} and \textsc{Centralized + ReconEval}).

The model learns $P$ and $Q$ such that $R \approx PQ^{\top}$ as discussed in \cref{sec:problem_setting}, with embedding dimensionality $K=50$. We report root-mean-square-error (RMSE) and rating prediction accuracy (rounding predicted ratings to the nearest integer, how often does the model predict the correct rating?).

We apply \fedrecon with local user embeddings $P_u$ and global item matrix $Q$. The dataset split function $S$ splits each user's data into half for the support and query sets (see \cref{sec:appendix_dataset_models_nwp} for a discussion of modifying this). $R$ and $U$ each perform up to 50 gradient descent steps. We tried smaller numbers of steps in \cref{fig:vary_steps}, and found that results mostly plateaued around 10 steps and stopped improving at 50 steps. We use a batch size of 5 for federated training. We grid over server learning rate $\lr_s \in [0.1, 0.5, 1.0]$, reconstruction learning rate $\lr_r \in [0.1, 0.5]$, and client update learning rate $\lr_u \in [0.1, 0.5]$. Note that $\lr_u$ corresponds to the client learning rate in the generalized \fedavg presented by \citet{reddi2020adaptive}. $\lr_r$ is newly introduced by our method, and we have found that setting it to the same as the client learning rate or slightly lower works well across tasks. We run 500 rounds of training with 100 clients randomly selected per round. For centralized baselines, we run 20 epochs of training with a batch size of 300. We report the configuration with best final validation performance for each setting. We rerun experiments 3x and report average metrics.

\subsection{Next Word Prediction}
\label{sec:appendix_dataset_models_nwp}
We use the federated Stack Overflow dataset introduced in \citet{TFFSO2019}. The dataset consists of public questions and answers from Stack Overflow, naturally partitioned into clients by posts from different users on the site. The dataset has 342,477 training clients with 135,818,730 examples and 38,758 held-out clients with 16,491,230 examples. The dataset is licensed under the Creative Commons Attribution-ShareAlike 3.0 Unported License \citep{stackoverflow_license}. 

We perform a autoregressive next word prediction task, predicting the next word in a sentence given the previous words. For ease of comparison, we process the data and use the model as described in (\citet{reddi2020adaptive} [Appendix C.4]); we restrict the vocabulary to one of [1000, 5000, 10000] most popular words and use padding and truncation to ensure each sentence has exactly 20 words. We also restrict each client to have at most 1000 sentences. We use the same LSTM with input embedding dimension 96 and output dimension 670, except we enable a variable number of OOV embeddings as motivated in \cref{sec:nwp_setup}. We compare across different numbers of OOV embeddings in \cref{fig:vary_oov} and observed that performance plateaus for all vocabulary sizes by 500 OOV, so we use this for experiments with multiple OOV embeddings. Just as \citet{reddi2020adaptive}, we use a local batch size of 16. We also use Yogi for the server update step. We report the top-1 accuracy, ignoring special tokens representing the out-of-vocabulary tokens, padding, and beginning/end of sentences.

We apply \fedrecon where the OOV embeddings are local and the rest of the model (including the core vocabulary embeddings) is global. The dataset split function $S$ splits each user's data in half by timestamp–we found that we achieved similar performance if the split was not half, as long as the split occurred by timestamp (otherwise, support and query examples are too similar and results are unrealistically inflated). $R$ and $U$ each perform up to 100 gradient steps. We tried smaller numbers of steps in \cref{fig:vary_steps}, and found that results mostly plateaued around 10 steps and stopped improving at 100 steps. We perform 2500 rounds of training with 200 clients randomly selected per round. We grid over server learning rate $\lr_s \in [0.01, 0.1, 0.3]$, reconstruction learning rate $\lr_r \in [0.1, 0.3]$, and client update learning rate $\lr_u \in [0.1, 0.3]$, similar to the grid used by \citet{reddi2020adaptive} and applying the heuristic that reconstruction learning rate can be set to about the same as client learning rate. Where applicable, we use the same hyperparameters and learning rate grids for the \fedavg baselines. We rerun experiments 3x and report average held-out accuracy for the best configuration for each setting.

\section{Additional Empirical Results}
\label{sec:appendix_empirical}

In \cref{tab:movielens_appendix_results,tab:so_nwp_appendix_results} we present experimental results for \fedrecon on matrix factorization and next word prediction tasks, in addition to the key results discussed in \cref{sec:experiment_results}. In these tables, we include some previously discussed results for ease of comparison. We call out interesting comparisons and additional figures in the discussion below.

\begin{table}[t]
\caption{Movielens matrix factorization root-mean-square-error (lower is better) and rating prediction accuracy (higher is better). \textsc{Standard Eval} is on seen users, \reconeval is on held-out users. Results within 2\% of best for each metric are in bold.}
\label{tab:movielens_appendix_results}
\vskip 0.15in
\begin{center}
\begin{small}
\begin{sc}
\begin{tabular}{lccr}
\toprule
 & RMSE $\downarrow$ & Accuracy $\uparrow$  \\
\midrule
Centralized + Standard Eval    & .923 & 43.2 \\
Centralized + Standard Eval (Unseen)   & 3.80 & 0.0 \\
Centralized + ReconEval   & 1.36 & 40.8 \\
\fedrecon (Ours)   & .907 & \textbf{43.3} \\
\fedrecon (No Split)   & .912 & 42.5 \\
\fedrecon (Joint Training)   & .915 & 42.1 \\
\fedrecon (Adagrad)   & \textbf{.883} & \textbf{44.1} \\

\bottomrule
\end{tabular}
\end{sc}
\end{small}
\end{center}
\vskip -0.1in
\end{table}

\begin{table}[t]
\caption{Stack Overflow next word prediction accuracy and communication per round, per client. \fedyogi and \textsc{OOV/Full Finetuning} require communication of all model parameters, \fedrecon does not (see \cref{fig:communication_plot}). Results within 2\% of best for each vocabulary size are in bold.}
\label{tab:so_nwp_appendix_results}
\begin{center}
\begin{small}
\begin{sc}
\begin{tabular}{lcccr}
\toprule
Vocab. Size & 1K & 5K & 10K  & Communication \\
\midrule
\fedrecon (1 OOV)   & 24.1 & 26.2 & 26.4  &  $2|g|$ \\
\fedrecon (500 OOV)   & \textbf{29.6} & \textbf{28.1} & \textbf{27.7}  &  $2|g|$  \\
\fedrecon (500 OOV, No Split)  & 28.8 & \textbf{28.1} & \textbf{27.7}  &  $2|g|$ \\
\fedrecon (500 OOV, Joint Training)  & \textbf{29.3} & \textbf{27.9} & \textbf{27.7}  &  $2|g|$ \\

\bottomrule
\end{tabular}
\end{sc}
\end{small}
\end{center}
\vskip -0.2in
\end{table}

\begin{figure}[t]
\vskip 0.0in
\centering
    \includegraphics[width=.5\columnwidth]{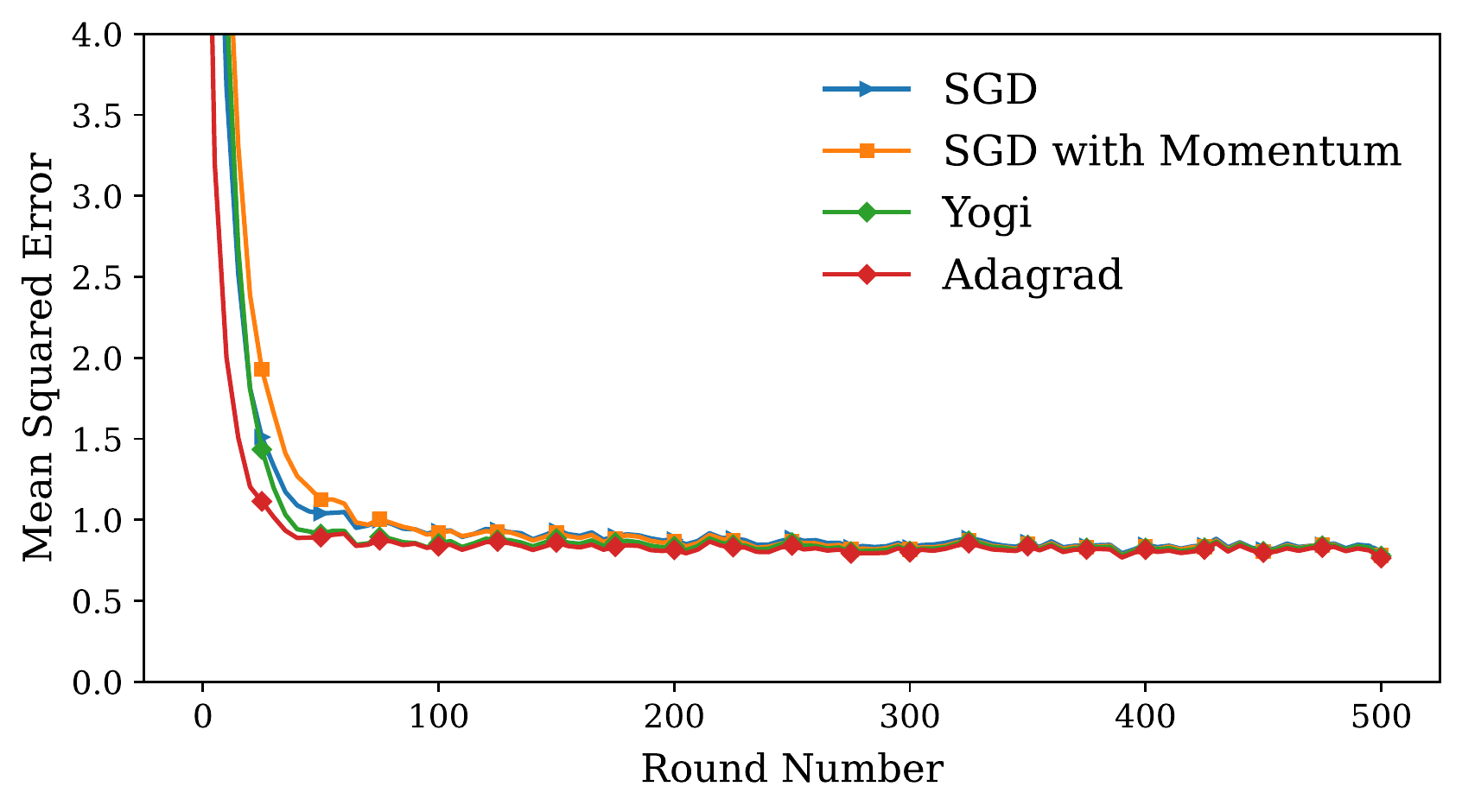}
    \vskip -0.1in
    \caption{MovieLens matrix factorization mean squared error (MSE) loss for \fedrecon across different server optimizers.}
\label{fig:movielens_optimizers}

\vskip -0.15in
\end{figure}

\begin{figure}[t]
\vskip 0.0in
\centering
    \includegraphics[width=.5\columnwidth]{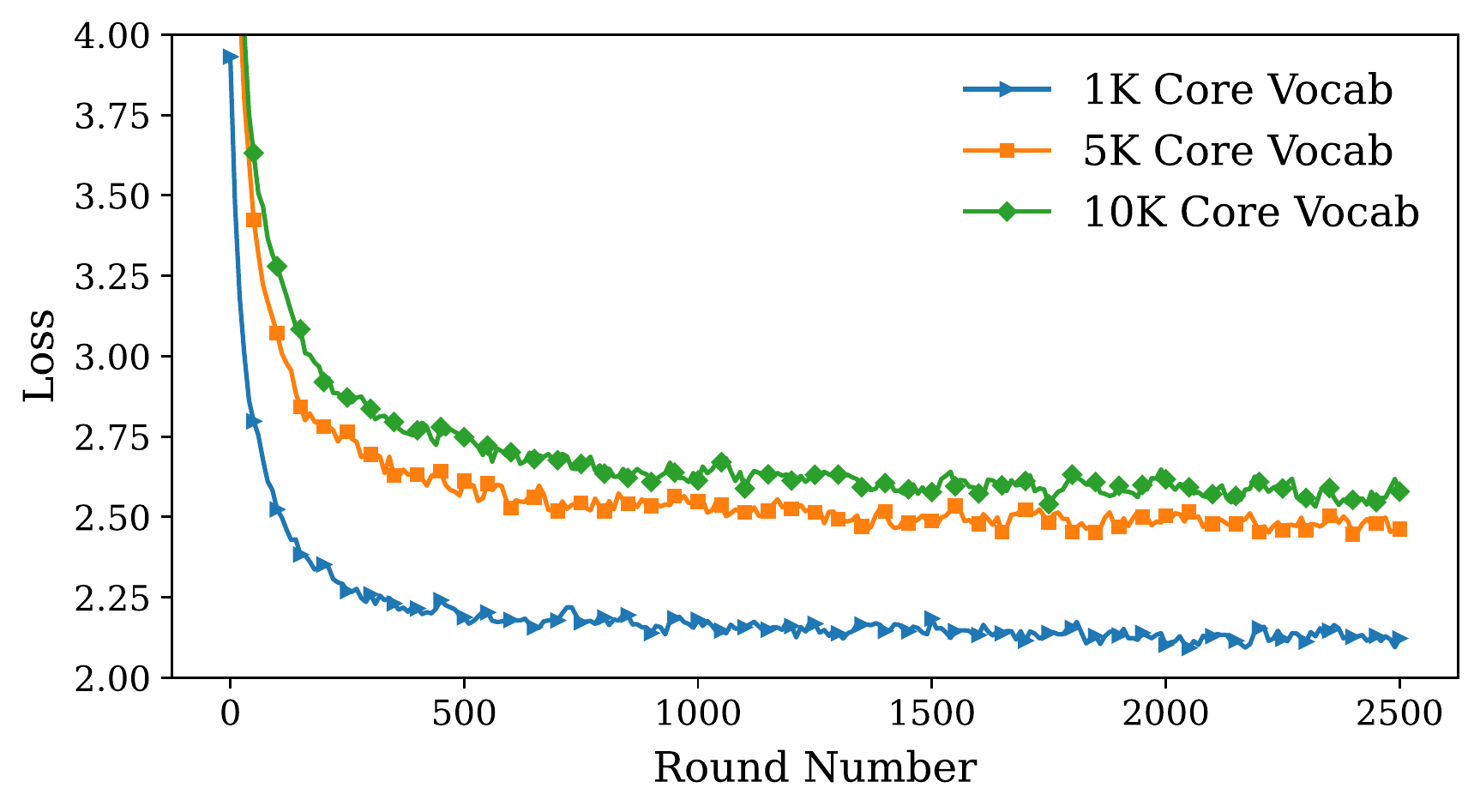}
    \vskip -0.1in
    \caption{Stack Overflow next word prediction cross-entropy loss for \fedrecon with varying core vocabulary size, with 500 out-of-vocabulary embeddings.}
\label{fig:nwp_losses}

\vskip -0.15in
\end{figure}

\begin{figure}[t]
\vskip 0.0in
\centering
    \includegraphics[width=.5\columnwidth]{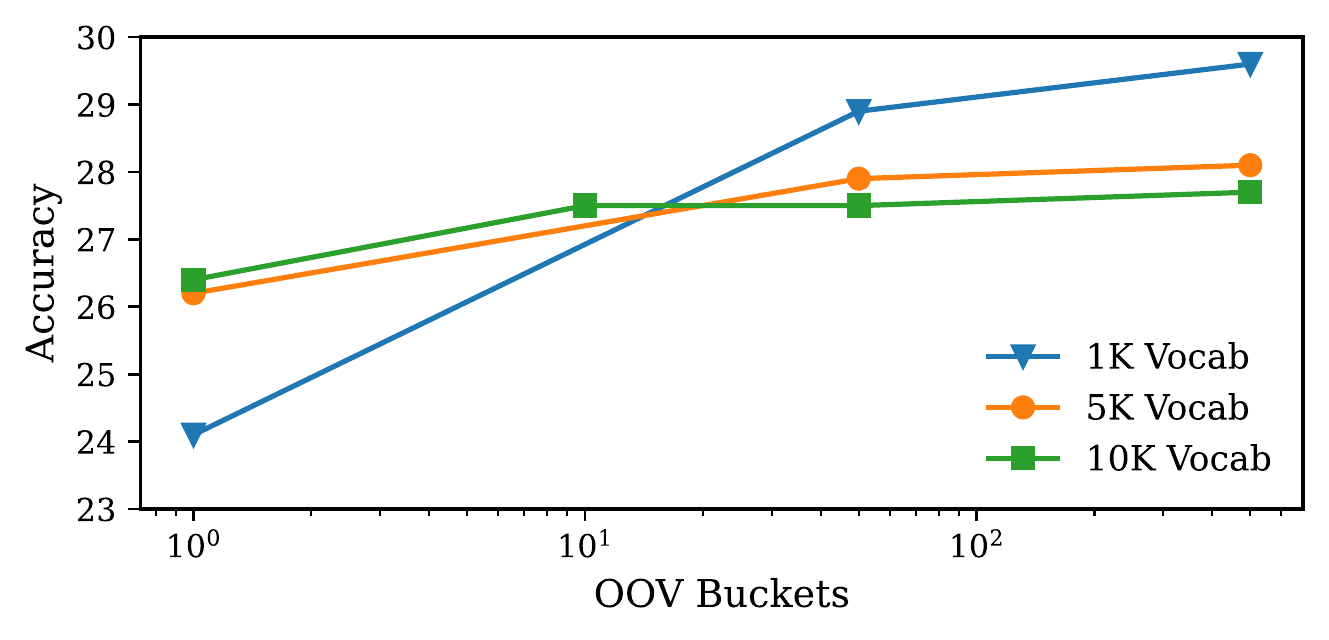}
    \vskip -0.1in
    \caption{Stack Overflow next word prediction accuracy for \fedrecon with varying numbers of out-of-vocabulary buckets for each core vocabulary size.}
\label{fig:vary_oov}

\vskip -0.15in
\end{figure}

\textbf{Sharing Support and Query Data: } If clients have very limited data (\eg some Stack Overflow clients have just one example) partitioning the data into disjoint support and query sets may be undesirable. As an ablation, in \cref{tab:movielens_appendix_results,tab:so_nwp_appendix_results} we evaluate \textsc{FedRecon (No Split)}, where the full client dataset is used for both support and query (keeping \reconeval the same for fairness); we see that we can relax the requirement for these sets to be disjoint with minimal drop in performance across MovieLens and Stack Overflow. We make use of this in our real-world deployment described in \cref{sec:deployment}, where it also helps combat data sparsity.

\textbf{Joint Training after Reconstruction: } \fedrecon involves alternating between (1) training local parameters with global parameters frozen during reconstruction and (2) training global parameters with local parameters frozen after reconstruction. We also tried joint training after reconstruction, where we update local and global parameters concurrently after reconstruction. In \cref{tab:movielens_appendix_results,tab:so_nwp_appendix_results} we evaluate \textsc{FedRecon (Joint Training)}. We see similar but slightly degraded performance in this setting. This suggests that freezing local parameters during (2) is useful for ensuring that global parameters are updated significantly (otherwise training does not progress across rounds since local parameters are not aggregated).

\textbf{Adaptive Optimizers: } \fedrecon treats aggregated global parameter updates as an "antigradient" that can be input into different server optimizers, building off of \citet{reddi2020adaptive}. In \cref{fig:movielens_optimizers} we observe that we can apply different server optimizers to the MovieLens matrix factorization task and loss converges well. In \cref{tab:movielens_appendix_results} we report an improved result for \fedrecon using \textsc{Adagrad}. All Stack Overflow results use Yogi as the server optimizer for consistency with \citet{reddi2020adaptive}–we also observed this improved performance over SGD. These results indicate that \fedrecon can be profitably combined with other advances in federated optimization.

\textbf{Stack Overflow Vocabulary and OOV Sizes: } In \cref{fig:nwp_losses} we plot loss over rounds for different core vocabulary sizes for the Stack Overflow next word prediction task (applying \fedrecon with 500 OOV embeddings fixed). We observe that losses are lower as core vocabulary coverage decreases, consistent with the hypothesis posited in \cref{sec:experiment_results} that lowering the size of the core vocabulary provides more "training data" for the local OOV embeddings, which improves local personalization performance (note that the same trend does not occur for the \fedyogi result from \cref{tab:so_nwp_baselines}).

In \cref{fig:vary_oov} we see that adding OOV buckets improves accuracy the most for a smaller core vocabulary, as expected. For all vocabulary sizes, performance plateaus around 500 OOV buckets.

\textbf{MovieLens Centralized Evaluation on Unseen Users: } In \cref{tab:movielens_appendix_results} we evaluate a standard server-trained matrix factorization model on unseen users in two ways: (1) standard evaluation, where we randomly initialize a local embedding for the user and compute metrics, and (2) \reconeval, where we split each user's data, reconstruct the user embedding using support data, and compute metrics on query data. Note that since unseen users do not have any trained user embeddings, these initialization strategies are needed for evaluation. (1) produces bad results as expected given the randomly initialized user embeddings, indicating that \reconeval is needed to produce reasonable results. As discussed in \cref{sec:experiment_results}, \fedrecon produces improved results with \reconeval on unseen users compared to centralized training because it trains global parameters conducive to reconstruction.

\section{Privacy Implications and Limitations}
\label{sec:appendix_privacy_implications}

\fedrecon enables clients to learn models without sending privacy-sensitive parameters to a central server, as in the matrix factorization application. More broadly, even for use-cases without user-specific parameters, our method also provides a practical alternative to centralized training (and centralized data collection) and fully global federated learning, with some of the benefits of fully local learning. We hope that our contributions push future machine learning applications towards requiring less centralized data collection and less communication of privacy-sensitive personal information; this is why we have open-sourced the code framework used to deploy \fedrecon in a large-scale real-world federated learning application (see \cref{sec:deployment}), making practical partially local federated learning widely available.

However, like other federated learning algorithms, the method involves clients sending gradient updates to be aggregated on a central server. Several works have shown that this can lead to leakage of client information to a curious server \citep{zhu2020deep,geiping2020inverting,wei2020framework,yin2021see}. These attacks are generally most successful in simple cases, \eg if each client has just one training example used to derive their gradient and performs one gradient step. Still, significant leakage and even reconstruction of training data has been observed in more realistic settings \citep{geiping2020inverting,yin2021see}.

Vanilla \fedrecon may naturally provide a degree of protection against gradient leakage attacks. Since only a subset of model parameters' updates are communicated to the server, and these updates are directly calculated using only the query subset of the data, the attacks introduced in previous work may not be applicable. However, further attacks could be developed and vanilla \fedrecon (as well as other federated learning algorithms) provides no formal guarantee that gradients for global parameters will not leak information. Thus, in the most privacy-sensitive applications, \fedrecon may be augmented with differential privacy \citep{dwork2014algorithmic} or secure aggregation \citep{bonawitz2017practical} applied to global parameter updates, to provide provable guarantees about the information the server receives. Below we describe why applying differential privacy may be particularly promising for \fedrecon.

\subsection{Application to Differential Privacy}
\label{sec:application_to_dp}
Federated learning has provided a natural application for differentially private computing \citep{dwork2014algorithmic,geyer2017differentially}. Differential privacy is typically applied in (federated) machine learning via variants of \dpsgd, as outlined by \citet{abadi2016deep} and \citet{mcmahan2017learning}. This algorithm clips and noises gradients in the spherical geometry of $\ell^2$ at each step of model training, introducing a dependence on model dimensionality. For an iterative procedure whose intermediate results are vectors in $\mathbb{R}^d$, each iteration must add noise at the scale of $\sqrt{d}$; this $\sqrt{d}$ subsequently appears as a steady-state risk in regret-based analysis of differentially private optimization; see \eg \citet{DBLP:journals/corr/BassilyST14}, \citet{DBLP:journals/corr/0001KCJN16}, \citet{DBLP:journals/corr/abs-1908-09970}. 

\fedrecon provides a natural parameter to improve the performance of differentially private federated training: the dimensionality of the global parameters. By splitting the model into local and global portions, \fedrecon reduces the dimensionality of aggregated parameters, reducing the variance of the noise which must be added to ensure user-level differential privacy \citep{mcmahan2017learning}. We emphasize that this is a systems-focused application of \fedrecon, rather than a fundamentally new differential privacy algorithm; tuning the number of global parameters composes rather than competes with algorithmic advances in differential privacy. We believe this is an interesting area for future work.

\end{document}